\documentclass[preprint,12pt]{elsarticle}

\usepackage{amssymb,multirow}
\usepackage{array} % 引入 array 包
\usepackage{amsmath}
\DeclareUnicodeCharacter{266B}{~} % 266B是音乐符号的Unicode编码
\usepackage{csquotes}
\usepackage{CJKutf8}
\usepackage{booktabs}
\usepackage{footnote}
\journal{Nuclear Physics B}

\begin{document}

\begin{frontmatter}

%% Title, authors and addresses

%% use the tnoteref command within \title for footnotes;
%% use the tnotetext command for theassociated footnote;
%% use the fnref command within \author or \affiliation for footnotes;
%% use the fntext command for theassociated footnote;
%% use the corref command within \author for corresponding author footnotes;
%% use the cortext command for theassociated footnote;
%% use the ead command for the email address,
%% and the form \ead[url] for the home page:
%% \title{Title\tnoteref{label1}}
%% \tnotetext[label1]{}
%% \author{Name\corref{cor1}\fnref{label2}}
%% \ead{email address}
%% \ead[url]{home page}
%% \fntext[label2]{}
%% \cortext[cor1]{}
%% \affiliation{organization={},
%%             addressline={},
%%             city={},
%%             postcode={},
%%             state={},
%%             country={}}
%% \fntext[label3]{}

\title{QCG-Rerank: Chunks Graph Rerank with Query Expansion in Retrieval-Augmented LLMs for Tourism Domain}

%% use optional labels to link authors explicitly to addresses:
%% \author[label1,label2]{}
%% \affiliation[label1]{organization={},
%%             addressline={},
%%             city={},
%%             postcode={},
%%             state={},
%%             country={}}
%%
%% \affiliation[label2]{organization={},
%%             addressline={},
%%             city={},
%%             postcode={},
%%             state={},
%%             country={}}

\author{Qikai Wei{$^{*,a,b}$}, Mingzhi Yang$^{*,c}$, Chunlong Han$^{a,b}$, Jingfu Wei$^{a,b}$, \\Minghao Zhang$^{a,b}$, Feifei Shi$^{a,b}$, Huansheng Ning{$^{\dagger,a,b}$}} %% Author name
%% Author affiliation
\affiliation{organization={School of Computer and Communication Engineering, University of Science and Technology Beijing, Beijing 100083, China},%Department and Organization
            }
\affiliation{organization={Beijing Engineering Research Center for Cyberspace Data Analysis and Applications, Beijing 100083, China},%Department and Organization
            }
\affiliation{organization={Guangxi Tourism Development One-Click Tour Digital Cultural Tourism Industry Co.,Ltd, Guangxi 530012, China},%Department and Organization
            }

\affiliation{addressline={weiqikai@xs.ustb.edu.cn, asanseu@163.com, hanchunlong@xs.ustb.edu.cn, m202220911@xs.ustb.edu.cn, m202311507@xs.ustb.edu.cn, shifeifei@ustb.edu.cn}, ninghuansheng@ustb.edu.cn
            }

%% Abstract
\begin{abstract}
Retrieval-Augmented Generation (RAG) mitigates the issue of hallucination in Large Language Models (LLMs) by integrating information retrieval techniques. However, in the tourism domain, since the query is usually brief and the content in the database is diverse, existing RAG may contain a significant amount of irrelevant or contradictory information contents after retrieval. To address this challenge, we propose the QCG-Rerank model. This model first performs an initial retrieval to obtain candidate chunks and then enhances semantics by extracting critical information to expand the original query. Next, we utilize the expanded query and candidate chunks to calculate similarity scores as the initial transition probability and construct the chunks graph. Subsequently, We iteratively compute the transition probabilities based on an initial estimate until convergence. The chunks with the highest score are selected and input into the LLMs to generate responses. We evaluate the model on Cultour, IIRC, StrategyQA, HotpotQA, SQuAD, and MuSiQue datasets. The experimental results demonstrate the effectiveness and superiority of the QCG-Rerank method.
\renewcommand{\thefootnote}{\fnsymbol{footnote}}
\footnotetext[1]{These authors contributed equally to this work.}
\footnotetext[2]{Corresponding author}
\renewcommand{\thefootnote}{\arabic{footnote}}

\end{abstract}%

%%Graphical abstract
% \begin{graphicalabstract}
% %\includegraphics{grabs}
% \end{graphicalabstract}

% %%Research highlights
% \begin{highlights}
% \item We construct Cultour, a high-quality Chinese SFT dataset for tourism and culture. The dataset contains tourism knowledge base QA data, travelogues data, and tourism diversity QA data.

% \item We propose TourLLM, a Qwen-based model supervised-finetuned with Cultour in the tourism domain. 

% \item We introduce CRA, a novel metric for evaluating LLMs in the tourism domain considering consistency, readability, and availability.

% \item We assess TourLLM using both automated and human evaluations. The experimental results demonstrated the effectiveness of TourLLM.
% \end{highlights}

%% Keywords
\begin{keyword}
%% keywords here, in the form: keyword \sep keyword

%% PACS codes here, in the form: \PACS code \sep code

%% MSC codes here, in the form: \MSC code \sep code
%% or \MSC[2008] code \sep code (2000 is the default)
RAG\sep Query Expansion\sep Rerank \sep Tourism\sep LLMs
\end{keyword}

\end{frontmatter}

%% Add \usepackage{lineno} before \begin{document} and uncomment 
%% following line to enable line numbers
%% \linenumbers

%% main text
%%

%% Use \section commands to start a section
\section{Introduction}
Recently, LLMs, such as Llama\cite{llama}, ChatGPT\cite{gpt}, and ChatGLM\cite{glm}, have strong generative capabilities and are widely applied in question answering\cite{qa_kg} and code generation\cite{code_generation}\cite{code_generation_2}. LLMs are trained on large-scale corpora to learn deep semantic information, enabling them to generate corresponding responses based on queries. However, the generated responses sometimes contain inconsistencies with facts, leading to the \enquote{hallucinations} phenomenon\cite{hallucinations}. In the tourism domain, LLMs often produce outdated addresses and route information, making the hallucination phenomenon particularly obvious\cite{cultour}.

\begin{figure}[t]
	\begin{center}
		\includegraphics[scale=0.75]{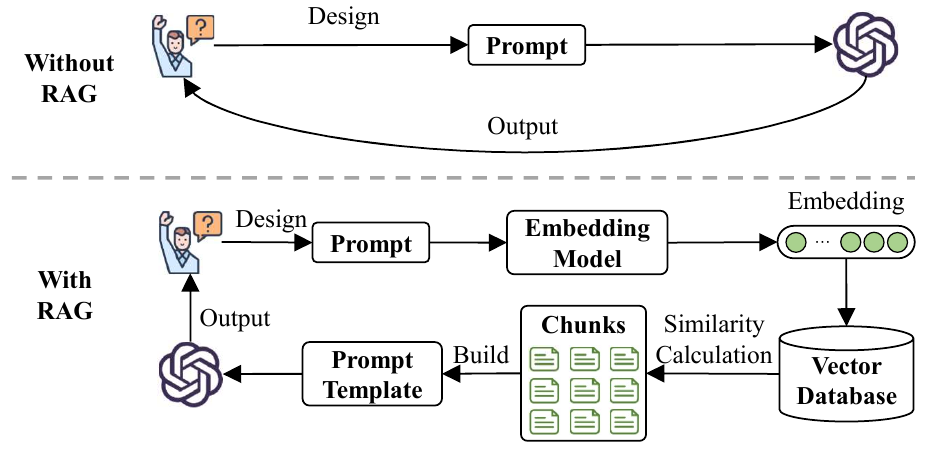}
		\caption{The process of without RAG and with RAG.}
		\label{fig:1}       % Give a unique label
	\end{center}
\end{figure}

To mitigate the hallucination issues of LLMs in the tourism domain, we use RAG to retrieve query-related contents in tourism and input them into LLMs. Subsequently, we use the matched contents to limit the scope of LLMs' responses, ensuring that the generated output is more consistent with the user's query\cite{gao2023retrieval}. Due to its outstanding performance, RAG has been widely applied in law\cite{lawyer_llama}\cite{Chatlaw} and medicine\cite{medical_graph_rag}. Fig. \ref{fig:1} illustrates the detailed processes LLMs with RAG and without RAG. Given the concise nature of queries in the travel domain and the inconsistent quality and length of candidate contents, similarity-based retrieval methods with travel documents often result in retrieving chunks irrelevant to or contradictory to the user's query. Consequently, this misalignment can cause LLMs to summarize incorrect information, adversely affecting the model's overall performance.

To address this issue, Wang et al.\cite{Query2doc} utilized LLMs to generate pseudo-documents for the original query and concatenated them to expand the semantic complexity of the query. The enhanced query is then used for retrieval to improve the matching accuracy of relevant documents. Given that the documents retrieved in the initial stage based on similarity may have low semantic relevance or contradict the original query, Xiao et al.\cite{C-pack} proposed a rerank method based on Bge-embedding. This method aims to refine the initial retrieval by adding an additional rerank step that leverages semantic understanding to identify the relevant contents. The reranked contents are then summarized using the capabilities of LLMs to provide more accurate responses. Youdao\cite{BCEmbedding} has developed bilingual and cross-lingual embedding (BCEmbedding) in English and Chinese. It includes two main components: the EmbeddingModel and the RerankerModel. The EmbeddingModel is designed to produce semantic vectors essential for enhancing semantic searches and facilitating question-answering processes. Meanwhile, the RerankerModel is adept at improving the quality of search outcomes and performing ranking tasks with precision.

To mitigate the hallucination problem in RAG in the travel domain, it is crucial to mine information deeply from brief queries and retrieve relevant contents accurately. In this paper, we propose \textbf{C}hunks \textbf{G}raph \textbf{Rerank} with \textbf{Q}uery Expansion(QCG-Rerank). We first perform an initial retrieval using the query to obtain candidate contents. Next, we extract critical information from the query and duplicate it to expand its semantic complexity to improve matching reliability. Given that the initial retrieval results may contain overlapping contents with minimal differences in similarity, our reranking stage involves calculating the semantic similarity between the updated query and the candidate chunks. These similarity scores serve as initial transition probabilities to construct a chunks graph. After several iterations, the transition probabilities stabilize, and we choose the top-ranked chunks to input into LLMs, which summarize the relevant contents and generate the final output. This approach ensures that the chunks input into the large model are highly relevant to the query, thereby enhancing the reliability of the responses generated by the LLMs in the travel domain. We evaluate the model on Cultour, IIRC, StrategyQA, HotpotQA, SQuAD, and MuSiQue datasets. The experimental results demonstrate the effectiveness and superiority of the QCG-Rerank method.

In summary, the contributions of our paper are as follows:

1.	We extract critical information from the query and duplicate it to enrich the semantic complexity of the original query, thereby enhancing the relevance of the retrieved contents.

2.	We propose a chunks graph reranking method that constructs a graph based on the similarities between the updated query and candidate chunks. Identifying the significant chunks within this graph enhances the accuracy of the reranking stage.

3.	We evaluate QCG-Rerank on Cultour, IIRC, StrategyQA, HotpotQA, SQuAD, and MuSiQue datasets. The experimental results highlight the superior performance and accuracy of QCG-Rerank.

\section{Related Work}
\subsection{RAG}

RAG excels at combining information retrieval with generative models, aiming to improve response accuracy and minimize hallucination issues in generated content.\cite{guu2020retrieval}. The RAG process consists of two primary stages. Initially, given a user's query, the retrieval component retrieves the most pertinent external contents. Following this, the generative component produces the final response based on retrieved contents.\cite{huang2024survey}. Due to its ability to alleviate hallucinations, RAG is extensively applied in open-domain question answering\cite{cultour}\cite{trivedi2022interleaving}\cite{liu2024eglr}, dialogue systems\cite{lawyer_llama}\cite{med_survey}\cite{Disc}\cite{bai2024infusing}, and code generation\cite{code_generation_2}\cite{yang2022ccgir}.

\subsection{RAG for LLMs}

Recently, RAG has been widely integrated into LLMs\cite{Chatlaw}. In this scenario, it retrieves relevant contents based on the query and then utilizes LLMs' summarization capabilities to respond to the query\cite{gao2023retrieval}\cite{peng2023check}\cite{xu2023search}. The essential aspect impacting the RAG capability in LLMs is the retrieval capability, which researchers have enhanced through optimizations in two key areas: (1) Embedding and (2) External document retrieval capability. For Embedding, a crucial component of RAG influences converting text into vectors, reducing the distance between texts in the vector space, thereby improving retrieval efficiency. Zhang et al.\cite{zhang2024multi} introduced LLM-Embedding, leveraging the ranking positions of N sample outputs from LLMs to refine feedback expectations and thus improve model reranking accuracy. Addressing the issue of low semantic representation quality and fragmented useful information due to using chunks as retrieval units, Luo et al.\cite{luo2024bge} developed Landmark Embedding, employing a sliding window non-blocking method to capture embeddings that preserve contextual consistency and enhance retrieval effectiveness. Given that current RAG retrieval tasks rely heavily on static rules, often focusing LLMs' attention on the final sentences or tags, Su et al.\cite{su2024dragin} introduced DRAGIN, a system designed to dynamically determine when and what to retrieve based on information needs, thereby expanding the scope and flexibility of the retrieval process.

For External document retrieval capability, researchers use query to find matching related contents. If the retrieved information is irrelevant to the query, the LLMs will output ambiguous answers, leading to hallucination problems. To mitigate this challenge, Wang et al.\cite{Query2doc} introduced query2doc, a method to create pseudo-documents using a few prompt-based LLMs. These pseudo-documents are then merged with the original query to form a more refined query, thereby combining LLMs' basic knowledge with external content. This approach enhances the precision and relevance of the final response. On the other hand, Shi et al.\cite{shi2024generate} proposed GenGround, which first uses the query input to LLMs to generate a basic answer. Subsequently, the retrieved content is used to refine and correct inaccuracies in the initial response. In addition, Ma et al.\cite{rrr} generated queries through LLMs and used web search engines to retrieve context, further improving the overall quality of the generated content.

To reduce the impact of noise information in contents on the final results, Xu et al.\cite{xu2024unsupervised} proposed a training method based on information refinement. This method optimizes RAG in an unsupervised manner and filters out noise information, enhancing the reliability and accuracy of the output results. On the other hand, Zhu et al.\cite{zhu2024information} introduced the information bottleneck theory into RAG. Their method maximizes the mutual information between compression and ground output while minimizing the information between compression and retrieved passage. Thereby improving the overall quality of the generated content.

To address the limitations of RAG in complex query contexts, Asai et al.\cite{asai2023self} proposed SELF-RAG, which introduced a self-reflection mechanism into the RAG field. This enables the model to evaluate the quality of generated text and use this feedback to optimize future retrieval and generation processes. Li et al.\cite{li2024role} argue that retrieving documents containing long-tail knowledge is crucial, especially when the input query involves niche or less common information. Focusing on such specific types of data is essential for improving the accuracy and relevance of the final results. Moreover, user queries are diverse, including simple questions and answers and complex queries that require single-step or multiple searches. Jeong et al.\cite{jeong2024adaptive} proposed an adaptive QA framework. This framework includes a classifier that dynamically selects the most suitable retrieval strategy based on the complexity of the user's query. This adaptive approach can effectively manage different types of queries and provide relevant responses.

While the methods mentioned above focus on various aspects of content retrieval and generation, they often overlook the importance of extracting critical information from the original query. In this paper, we extract critical information from the original query based on the prompt method and expand it to increase the semantic complexity. This can enhance the retrieved content's relevance to the query to improve the reranking stage's performance. To further optimize the final results, we construct a chunks graph with relevant chunks obtained from the initial retrieval. Next, we propose a chunks graph rerank algorithm to optimize the final result, select essential chunks from the chunks graph, and input them into LLMs to enhance the reliability of the final response.
\section{Our Approach}
In this section, we introduce the QCG-Rerank model in detail. The model consists of three parts, as shown in Fig. 2. First, we fine-tune the embedding model using tourism data to ensure it captures more semantic information relevant to the tourism domain. Then, we use the fine-tuned embedding to vectorize all documents and build a vector database. Next, we vectorize the query to retrieve relevant documents in the vector database (Section 3.1). Since the query is brief and contains limited semantic information, making retrieval more challenging, we use a prompt-based method to extract and splice critical information from the query. Additionally, we duplicate the query to enhance its semantic complexity during retrieval (Section 3.2). Finally, we use the retrieval results and the query integrating critical information to build a chunks graph. We use the chunks graph rerank method to filter out the top-ranked chunks in the chunks graph and input them into LLMs (Section 3.3). After the above steps, LLMs respond by combining the filtered chunks with the query.
\begin{figure}[t]
	\begin{center}
		\includegraphics[scale=0.65]{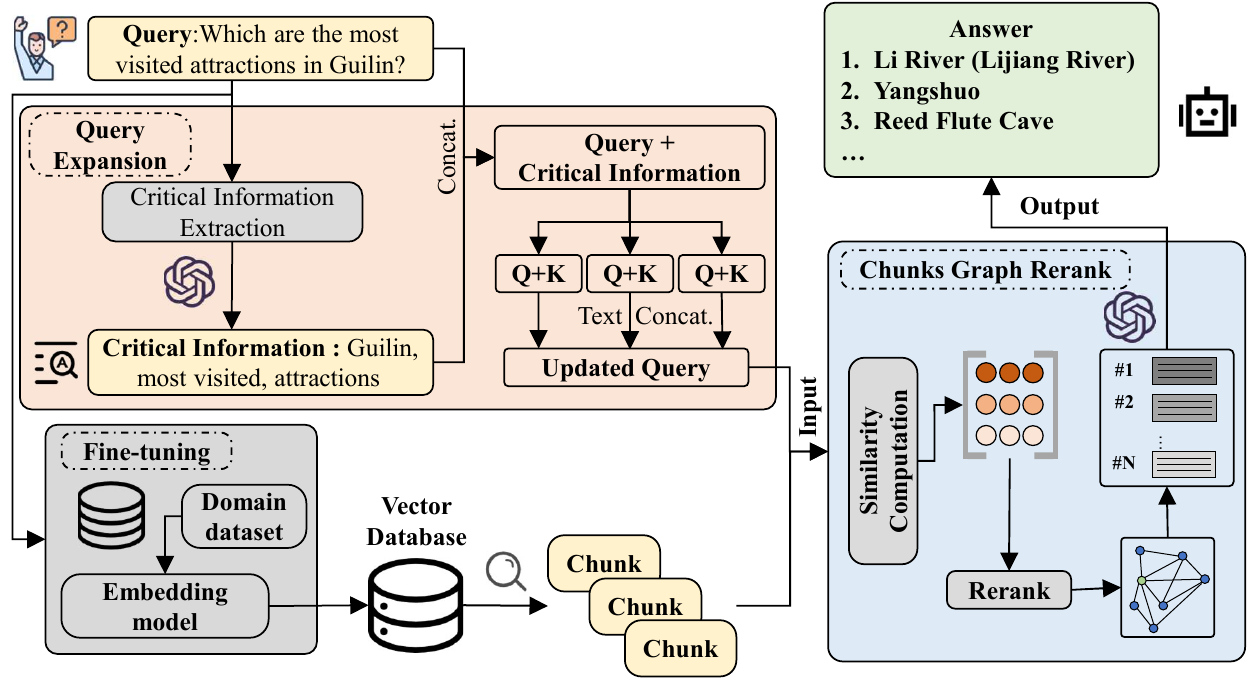}
		\caption{The overall framework of the QCG-Rerank model.}
		\label{fig:2}       % Give a unique label
	\end{center}
\end{figure}
\subsection{Fine-tuning embeddings}
Feng et al.\cite{feng2022novel} found that using domain data to fine-tune the embedding model can better capture basic semantic information. Inspired by Xiao et al.\cite{C-pack}, we construct a training corpus in the format of {\enquote{query}: str, \enquote{pos}: List[str], \enquote{neg}: List[str]}, where \enquote{pos} represents the correct answers and \enquote{neg} represents documents randomly selected from other answers. Then, We use them to fine-tune the embedding model, improving its domain adaptability and performance in capturing relevant semantic information.

After fine-tuning the embedding model, we convert each chunk $C_i \in Con$ into a corresponding vector $e_i \in \mathbb{R}^D$ using the tourism embedding model and save these vectors in the vector database. $D$ represents the dim of vector, $Con$ represents original contents. We then use the fine-tuned embedding to encode the query $q_o$ into the corresponding query vector $v_q \in \mathbb{R}^D$. Next, we calculate the similarity between $v_q$ and each chunk vector $e_i$ in the vector database to identify the top $N$ most similar chunks. The calculation process is shown in Eq. (\ref{eq:1}).
\begin{equation}
\text{Set} = \underset{N}{\text{Top}}\left( \frac{v_q \cdot C_i}{\|v_q\| \|C_i\|} \right)
\label{eq:1}
\end{equation}
Where Set represents the candidate contents related to $q_o$ obtained after retrieval.
\subsection{Critical Information Extraction}
\begin{figure}[t]
	\begin{center}
		\includegraphics[scale=0.75]{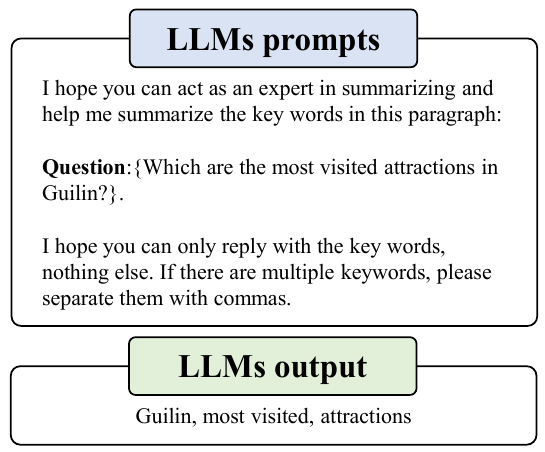}
		\caption{Prompt template for query expansion.}
		\label{fig:3}       % Give a unique label
	\end{center}
\end{figure}

In tourism, users' queries are often brief, such as \enquote{Which are the most visited attractions in Guilin?}. Meanwhile, the chunks in the database are generally longer, leading to matching texts that are not highly relevant or even contradictory to the query during retrieval. This affects the quality of the retrieved chunks and the output results of LLMs. To address this question, we design a prompt template to extract critical information from the given query. The template is shown in Fig. \ref{fig:3}.

The template involves assigning roles to LLMs and leveraging their inherent information extraction capabilities to distill the critical information from queries. Subsequently, the query $q_o$ and the extracted critical information $CR\_inf$ are concatenated to update query $q_{up}$. To ensure that the query $q_{up}$ aligns more closely with the chunks stored in the database within the vector space, following the approach by Wang et al.\cite{Query2doc}, we create the final query by duplicating several times. The concatenation procedure is illustrated in Eq. (\ref{eq:2}).
\begin{equation}
q_{up} = \text{concat} \left( (q + CR\_inf) \times n \right)
\label{eq:2}
\end{equation}
Where \enquote{concat} represents the concatenation function, $q_{up}$ represents the updated query that concatenates the query and the critical information, which is used as the query in the rerank process, and $n$ represents the number of duplication.

\subsection{Chunks Graph Rerank}
We proposed a chunks graph rerank method to address some documents in the recall results that have low relevance or conflicting chunks with the query. First, the recall results and the updated query $q_{up}$ form the chunks graph, and then we perform graph rerank to find chunks with higher scores in the chunk graph.

\begin{figure}[t]
	\begin{center}
		\includegraphics[scale=0.6]{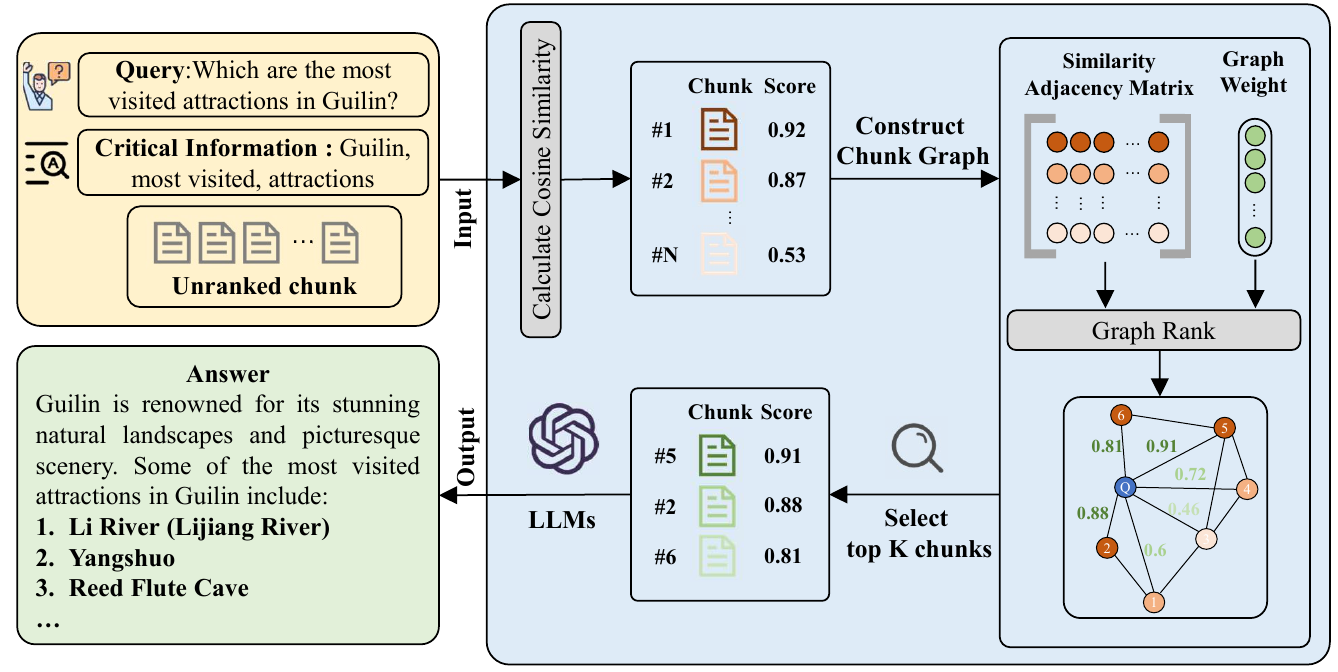}
		\caption{The overall framework of the Chunks Graph Rerank algorithm.}
		\label{fig:4}       % Give a unique label
	\end{center}
\end{figure}

Subsequently, we convert the query $q_{up}$ obtained in Section 3.2 into the corresponding query vector $v_{up}$ using the fine-tuned embedding model. Then, we use cosine similarity $sim_{up}$ to calculate $v_{up}$ and each candidate chunk vector to obtain the similarity score between each chunk and the updated query. The calculation process is shown in Eq. (\ref{eq:3}).
\begin{equation}
\text{sim}_{up} = \frac{v_{up} \cdot \text{Set}_i}{\|v_{up}\| \| \text{Set}_i \|}, \quad i \in \{1, \ldots, N\}
\label{eq:3}
\end{equation}

Inspired by TextRank\cite{Textrank}, we construct the chunks graph $G$ using the query $q_{up}$ and the candidate document Set. We then construct the similarity transfer matrix using the similarity scores. In graph $G$, only the positions with an interactive relationship with the query are assigned the corresponding similarity scores, while the relationships between contents are initialized to 0. Additionally, we initialize the weight of the graph as shown in Eq. (\ref{eq:4}).

\begin{equation}
S_i = 
\begin{cases} 
1, & \text{if } i = 0 \\
0, & \text{else}
\end{cases}, \quad i \in \{1, 2, \ldots, num, num + 1\}
\label{eq:4}
\end{equation}
Where $num$ represents the total number of chunks obtained by retrieve, and $S_i$ represents the weight of the graph $G$ node. It enhances the weight coefficient of the query in the rerank process.

We input the chunks graph $G$ and the initial weight $S_i$ into the chunks graph rank algorithm and use the query to guide the iterative process to select the chunks with higher scores in the graph. The iterative calculation process is shown in Eq. (\ref{eq:5}).

\begin{equation}
S^{(t+1)}(i) = (1 - d) S^{(t)}(i) + d \sum_{j \in \text{In}(i)} \frac{w_{ji}}{\sum_{k \in \text{Out}(j)} w_{jk}} S^{(t)}(j)
\label{eq:5}
\end{equation}
Where $S(i)$ represents the weight of node $i$, which is continuously updated with the number of iterations $t$, $d$ represents the damping coefficient, $In(i)$ represents all nodes connected to node $i$, $Out(j)$ represents all nodes that can be reached from node $j$, and $w_{ij}$ represents the correlation between node $i$ and node $j$.

\begin{figure}[t]
	\begin{center}
		\includegraphics[scale=0.85]{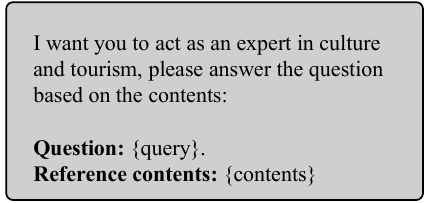}
		\caption{A template for answering questions based on contents.}
		\label{fig:5}       % Give a unique label
	\end{center}
\end{figure}

After the above operations, we can filter out the top $K$ chunks with the highest scores in the chunks graph and combine the query input into the designed template. Using the summarization ability of the LLMs, we can generate a response to the query. The calculation process is shown in Eq. (\ref{eq:6}):

\begin{equation}
\text{output} = \Theta(\psi(q_{up}, top_k(G)))
\label{eq:6}
\end{equation}
Where $\Theta$ represents LLMs, $\psi$ represents the template, and the designed template is shown in Fig. \ref{fig:5}. $top_k()$ represents sorting according to the weight information of $S(i)$ and returning the corresponding top $K$ chunks. Moreover, the output represents the response to the query after LLMs integrate the documents.

After the above process, we get the response of LLMs integration and chunks selected by QCG-rerank algorithm. The algorithm is as follows:

\begin{table}[h]
    \centering
    \footnotesize
    \label{tab:qcg_rerank_algorithm}
    \begin{tabular}{|p{13.5cm}|}
        \toprule
        \textbf{Input}  Query $q_o$, All documents chunks, Tourism dataset \\
        \textbf{Output}  LLMs response combined with all relevant chunks  \\
        \midrule
        1.  Begin \\
        2.  Fine-tune embedding using tourism dataset \\
        3.  Convert all contents into corresponding content vectors and build Vector DB \\
        4.  for $q_o$ in query sets: \\
        5.  \quad $Set$ $\leftarrow$ Utilize $q_o$ and all chunk vectors to compute the similarity based on Eq. (1) to get candidate chunks \\
        6.  \quad $CR\_inf$ $\leftarrow$ Extract critical information from $q_o$ with LLMs \\
        7.  \quad $q_{up}$ $\leftarrow$ Concatenate $CR\_inf$ with the $q_o$\\
        8.  \quad $v_{up}$ $\leftarrow$ Vectorize the updated query $q_{up}$ \\
        9.  \quad sim$_{up}$ $\leftarrow$ Use $q_{up}$ and top $N$ retrieved chunks to calculate the similarity based on Eq. (3)\\
        10.  \enspace\enspace $G$ $\leftarrow$ Use sim$_{up}$ to construct a chunks graph\\
        11.  \enspace\enspace $S_i$ $\leftarrow$ initialized based on Eq. (4) and iteratively updated based on Eq. (5)\\
        12.  \enspace\enspace Top $K$ rerank chunks $\leftarrow$ Select chunks with higher score \\
        13.  \enspace\enspace Response $\leftarrow$ Tnput top $K$ rerank chunks and $q_{up}$\\
        14.  End begin \\
        \bottomrule
    \end{tabular}
\end{table}

\section{Experimental}
In this section, we evaluate the effectiveness of our proposed QCG-Rerank method by comparing it with multiple models across several datasets. We provide a detailed analysis of the results, including an introduction to the datasets, baselines, hyperparameters, experimental settings, and evaluation metrics used.

\subsection{Datasets}
In this section, we introduce the six datasets used in this paper in detail: Cultour\cite{cultour}, IIRC\cite{iirc}, StrategyQA\cite{strategyqa}, HotpotQA\cite{hotpotqa}, SQuAD\cite{squad}, and MuSiQue\cite{musique}.

\textbf{Cultour} is a tourism dataset containing 12,000 question-answering examples. The data includes both self-collected tourism data and data automatically generated using LLMs. This dataset is used to evaluate the model's performance in the tourism field.

\textbf{IIRC} is an English dataset designed based on English Wikipedia paragraphs, comprising over 13,000 entries. Each question offers only fragmentary hints, with the full details dispersed across one or more associated articles. We utilize this dataset to assess the performance of model in reading comprehension tasks.

\textbf{StrategyQA} is an open-domain dataset featuring 2,780 examples. Each example includes a question and a supporting evidence paragraph. This dataset is used to evaluate the commonsense reasoning capabilities of models.

\textbf{HotpotQA} is a QA dataset derived from English Wikipedia, comprising approximately 113,000 questions. Each question requires finding answers by integrating information from paragraphs within two related articles. This dataset is used to assess the model’s ability to answer multi-step questions.

\textbf{SQuAD} is a reading comprehension dataset where the answer to each question is derived from a specific paragraph or span of text from a Wikipedia article. This dataset is used to evaluate the model’s performance in reading comprehension tasks.

\textbf{MuSiQue} is a multi-hop QA dataset where most questions require 2-4 reasoning steps to find the correct answer. This dataset is used to evaluate the model’s ability to perform multi-hop reasoning tasks.

\subsection{Baseline}
In this section, we introduce the baselines compared in this paper.

\textbf{Wo-RAG} is a model that uses LLMs to provide direct answers without searching for related documents.

\textbf{W-RAG} is a model that retrieves relevant documents in a corpus based on a query and then adds the retrieved documents directly to LLMs.

\textbf{BM25}\cite{bm25} is a statistical information retrieval algorithm that evaluates the relevance of a query to a document by calculating the term frequency and inverse document frequency of the query term in the document and obtains the final score through weighted summation.

\textbf{BM25L}\cite{bm25L} is an enhanced version of the BM25 algorithm that adjusts the inverse document frequency (IDF) weight and modifies the text length normalization factor. This adjustment reduces the algorithm’s bias towards shorter documents and improves its handling of longer documents, thereby enhancing the relevance and accuracy of information retrieval results.

\textbf{BGE-Rerank}\cite{C-pack} uses the query to perform two retrievals in an external database. The first retrieval employs a semantic similarity calculation method. The second retrieval uses the Bge-Rerank model to fine-tune and sort these results based on semantic relevance, thereby inputting more relevant documents into LLMs.

\textbf{BCE-Rerank}\cite{BCEmbedding} is a bilingual and cross-lingual semantic representation algorithm developed by NetEase Youdao. It excels at optimizing semantic search results and improving the ranking of semantically related items.

\begin{table}[t]
    \centering
    \setlength{\tabcolsep}{4pt} % 减小列间距
    \scriptsize
    
    \begin{tabular}{cccccc}
        \toprule
        Datasets & \#train & \#test & Learning rate & K & Train epoch  \\
        \midrule
        Cultour & 9,987 & 2,498 & 1e-5 & 3 & 5.0  \\
        IIRC & 4,906 & 593 & 1e-5 & 2 & 5.0  \\
        StrategyQA & 1,859 & 431 & 1e-5 & 5 & 5.0  \\
        HotpotQA & 5,924 & 1,481 & 1e-5 & 2 & 5.0 \\
        SQuAD & 8,456 & 2,114 & 1e-5 & 1 & 5.0  \\
        MuSiQue & 1,934 & 483 & 1e-5 & 1 & 5.0  \\
        \bottomrule
    \end{tabular}
    \caption{Details of the datasets and corresponding hyperparameters.}
    \label{tab:1}
\end{table}
\subsection{Experiment settings}
For a fair experiment, we use Qwen-2-7B-Instruct as the base LLMs and set the temperature to 0.0. In fine-tuning embedding, we evaluate models using both Bge-base and Bge-large embedding. During fine-tuning of the embedding, we set the training epochs to 5.0, the maximum query length to 64, the maximum passage length to 256, and the learning rate to 1e-5. Additionally, we set the number of duplications of critical information K to the range \{1, 2, 3, 4, 5\} based on the characteristics of each dataset. For the IIRC dataset, we use 4,906 examples as the training set and 593 examples as the test set. For the Culture and StrategyQA datasets, we split the training set into a 2:8 ratio, using 80\% of the data for fine-tuning the embeddings and 20\% for testing. For the HotpotQA, SQuAD, and MuSiQue datasets, we split the test set into a 2:8 ratio, using 80\% of the examples to fine-tune the embeddings and 20\% for testing. If the test set contains more than 1,000 data points, we randomly select 1,000 samples for testing. The detailed configurations are summarized in Table \ref{tab:1}. All experiments are conducted on 2 NVIDIA GeForce RTX 3090 GPUs.

\subsection{Evaluatation metrics}
In evaluating the model, due to the unique nature of the Strategy dataset, where answers are binary (\enquote{true} or \enquote{false}), we use Accuracy to measure the accuracy of the model's predictions on this dataset. For the other five datasets, we employ ROUGE\cite{rouge}, BLEU\cite{bleu}, and METEOR\cite{meteor}. ROUGE (Recall-Oriented Understudy for Gisting Evaluation) calculates the similarity between the automatically generated results and the manually generated reference answers. BLEU (Bilingual Evaluation Understudy): measures the accuracy of the results by comparing the n-gram overlap between the candidate results and the reference results, which is particularly useful when there are multiple correct answers. METEOR (Metric for Evaluation of Translation with Explicit Ordering) evaluates the quality of the model's output by considering factors such as synonym matching, stem matching, and word order matching, making it a more comprehensive evaluation metric than BLEU.

\section{Experiment Results}
\begin{table}[t]
    \centering
    \scriptsize
    
    \label{tab:performance_comparison}
    \setlength{\tabcolsep}{4pt} % 减小列间距
    \begin{tabular}{cc|cccc|cccc|c}
        \toprule
        \multirow{2}{*}{Models}&\multirow{2}{*}{Embedding}
        
        & \multicolumn{4}{c|}{Cultour} & \multicolumn{4}{c|}{IIRC} & StrategyQA \\
        \cmidrule(lr){3-6} \cmidrule(lr){7-10} \cmidrule(lr){11-11}
         && R-1 & R-L & B-1 & Met. & R-1 & R-L & B-1 & Met. & Acc \\
        \midrule
         Wo-RAG &Bge-base& 27.08 & 18.92 & 36.72 & 20.33 & 7.51 & 7.36 & 16.67 & 12.6 & 62.88 \\
         W-RAG &Bge-base &62.80 & \underline{57.92} & 52.83 & \underline{49.10} & 23.22 & 23.08 & 25.50 & 27.12 & 75.41 \\
         Bge-Rerank&Bge-base & 62.38 & 57.40 & 52.85 & 48.73 & 24.17 & 23.93 & 26.59 & 27.83 & \underline{77.03} \\
         BCE-rerank &Bge-base& \underline{62.84} & 57.93 & 52.73 & 48.90 & \underline{25.03} & \underline{24.90} & \underline{26.98} & \underline{28.87} & 76.80 \\
         BM25 &Bge-base &61.78 & 56.84 & \underline{53.24} & 47.92 & 23.39 & 24.22 & 26.84 & 28.39 & 76.10 \\
         BM25L &Bge-base &61.94 & 56.95 & 53.13 & 47.98 & 24.67 & 24.48 & 26.97 & 28.41 & 76.57 \\
         QCG-rerank &Bge-base& \textbf{64.13} & \textbf{59.55} & \textbf{53.51} & \textbf{50.08} & \textbf{27.18} & \textbf{26.93} & \textbf{29.21} & \textbf{30.33} & \textbf{78.89} \\
        \midrule
         Wo-RAG &Bge-large &27.08 & 18.93 & 36.71 & 20.34 & 7.48 & 7.33 & 16.76 & 12.61 & 62.88 \\
         W-RAG &Bge-large &62.27 & 57.53 & 52.38 & 48.62 & 25.56 & 25.43 & 27.61 & 28.36 & 76.10 \\
         Bge-Rerank &Bge-large &62.37 & 57.46 & 52.77 & 48.77 & 25.42 & 25.24 & 27.35 & 29.05 & 75.41 \\
         BCE-rerank &Bge-large &\underline{62.92} & \underline{58.01} & 52.97 & \underline{48.98} & \underline{26.56} & \underline{26.39} & 27.52 & \underline{29.98} & 77.02 \\
         BM25 &Bge-large &62.14 & 57.33 & 53.66 & 48.27 & 25.43 & 25.23 & 27.99 & 29.28 & \underline{77.26} \\
         BM25L &Bge-large &62.10 & 57.23 & \underline{53.51} & 48.17 & 25.70 & 25.53 & \underline{28.06} & 29.50 & 77.03 \\
         QCG-rerank &Bge-large &\textbf{64.69} & \textbf{60.12} & \textbf{54.41} & \textbf{50.49} & \textbf{28.03} & \textbf{27.89} & \textbf{28.86} & \textbf{31.26} & \textbf{77.49} \\
        \bottomrule
    \end{tabular}
    \caption{The overall experimental results of QCG-rerank and other baselines on Cultour, IIRC, StrategyQA datasets. The best results are in bold, and the second-best results are underlined.}
\end{table}
In this section, We evaluated the performance of the QCG-Rerank model and the baseline on the Cultour\cite{cultour}, IIRC\cite{iirc}, StrategyQA\cite{strategyqa}, HotpotQA\cite{hotpotqa}, SQuAD\cite{squad}, and MuSiQue\cite{musique} datasets using ROUGE-1(R-1), ROUGE-L(R-L), BLEU-1(B-1), METEOR(Met.) and Accuracy(Acc) metrics. The experimental results are shown in Table 2 and Table 3.

\begin{table}[t]
    \centering
    \setlength{\tabcolsep}{1pt} % 减小列间距
    \scriptsize
    
    \label{tab:my_label}
    \begin{tabular}{cc|cccc|cccc|cccc}
        \toprule
        \multirow{2}{*}{Models}&\multirow{2}{*}{Embedding}
        & \multicolumn{4}{c}{HotpotQA} & \multicolumn{4}{c}{MuSiQue}& \multicolumn{4}{c}{SQuAD} \\
        \cmidrule(lr){3-6} \cmidrule(lr){7-10}\cmidrule(lr){11-14}
        && R-1 & R-L & B-1 & Met. & R-1 & R-L & B-1 & Met. & R-1 & R-L & B-1 & Met. \\
        \midrule
        Wo-RAG &Bge-base& 15.92 & 15.75 & 29.15 & 20.71 & 5.00 & 4.90 & 15.00 & 12.25 & 9.26& 8.96	&17.59	&13.96 \\
        W-RAG &Bge-base& 34.37 & 34.24 & \underline{32.75} & 37.25 & 16.63 & 16.36 & 17.17 & 23.04 & 24.57&13.07	&22.69&	28.67 \\
        Bge-Rerank &Bge-base& 34.57 & 34.47 & 32.57 & \underline{37.85} & 18.95 & 18.62 & 18.55 & 24.83 & 26.08&13.50&24.29&	29.60 \\
        BCE-reranker &Bge-base& 34.12 & 34.02 & 32.14 & 37.31 & 17.03 & 16.68 & 17.06 & 23.21 & \underline{28.65}&\underline{28.32}& \underline{26.00}	&\underline{31.98} \\
        BM25 &Bge-base& 34.52 & 34.53 & 32.56 & 37.08 & \underline{19.58} & \underline{19.33} & \underline{19.59} & \underline{25.55} & 26.68&26.56	&24.51&	30.62 \\
        BM25L &Bge-base& \underline{34.80} & \underline{34.68} & 32.70 & 37.21 & 19.27 & 19.04 & 19.30 & 25.16 & 27.08&26.79	&24.48	&30.67 \\
        QCG-rerank &Bge-base& \textbf{34.86} & \textbf{34.76} & \textbf{32.82} & \textbf{38.31} & \textbf{20.78} & \textbf{20.38} & \textbf{20.85} & \textbf{26.70} & \textbf{30.27}&\textbf{30.05}&\textbf{26.92}&	\textbf{33.40} \\
        \midrule
        Wo-RAG &Bge-large& 15.95 & 15.78 & 29.24 & 20.69 & 5.06 & 4.96 & 15.05 & 12.33 & 9.19&8.91	&17.46&	13.92 \\
        W-RAG &Bge-large& 33.50 & 33.36 & 32.47 & 36.82 & 18.66 & 18.38 & 19.40 & 24.00 & 25.64&25.36	&24.52&	29.22 \\
        Bge-Rerank &Bge-large& \underline{35.15} & \underline{35.00} & \textbf{33.56} & \underline{37.89} & 19.36 & 19.00 & 19.33 & \underline{25.08} & 26.90&26.70&	24.99&	30.33 \\
        BCE-reranker &Bge-large& 34.97 & 34.87 & \underline{33.45} & 37.43 & \underline{19.38} & \underline{19.04} & \textbf{20.05} & 24.80 & \underline{29.00} &\underline{28.75}&	\underline{26.46}&	\underline{31.83}\\
        BM25 &Bge-large& 34.22 & 34.09 & 32.2 & 36.92 & 17.20 & 16.95 & 18.02 & 23.45 & 28.25&27.96	&26.33&	31.26 \\
        BM25L &Bge-large& 34.96 & 34.24 & 32.1 & 36.69 & 17.24 & 16.97 & 18.11 & 23.72 & 28.36&28.05	&25.92&	31.29 \\
        QCG-rerank &Bge-large& \textbf{35.44} & \textbf{35.3} & 32.69 & \textbf{38.61} & \textbf{20.03} & \textbf{19.73} & \underline{19.53} & \textbf{26.31} & \textbf{31.39}&\textbf{31.19}&	\textbf{27.42}&	\textbf{34.81} \\
        \bottomrule
    \end{tabular}
    \caption{The overall experimental results of QCG-rerankand other baselines on HotpotQA, MuSiQue, SQuAD datasets. The best results are in bold, and the second-best results are underlined.}
\end{table}

The experimental results in Table 2 and Table 3 demonstrate the improvements of the QCG-rerank method. Specifically, for the Chinese dataset Cultour, QCG-rerank shows a notable enhancement, particularly in the R-1 metric when utilizing the Bge\_large embedding, achieving a 2.32\% increase over Bge\_rerank. QCG-rerank also exhibits significant advancements in the English datasets IIRC, Strategy, and SQuAD. Notably, our approach achieves superior performance across most metrics on the HotpotQA and MuSiQue datasets. Additionally, using the Bge\_large embedding gets the best results on most of the test datasets. Experimental results demonstrate that the QCG-rerank method is effective in expanding the semantic information of the query, selecting relevant documents through chunks graph rerank, and inputting them into LLMs to generate the final response.

\subsection{Ablation experiment}
To assess the influence of various modules of QCG-rerank, we designed ablation experiments on the Chinese tourism dataset Cultour and the English dataset IIRC. Ablation experiments assessed the impact of two key components: duplicate critical information and chunk graph rerank. The detailed results are shown in Table 4.

\begin{table}[h]
    \centering
    \setlength{\tabcolsep}{2pt} % 减小列间距
    \scriptsize
    
    \label{tab:comparison}
    \begin{tabular}{c|cccc|cccc}
        \toprule
        \multirow{2}{*}{Models}
        & \multicolumn{4}{c}{Cultour} & \multicolumn{4}{c}{IIRC} \\
        \cmidrule(lr){2-5} \cmidrule(lr){6-9}
        & R-1 & R-L & B-1 & Met. & R-1 & R-L & B-1 & Met. \\
        \midrule
        QCG-rerank & \textbf{64.69} & \textbf{60.12} & \textbf{54.41} & \textbf{50.49} & \textbf{28.03} & \textbf{27.89} & \textbf{28.86} & \textbf{31.26} \\
        Wo-CG & 62.36 & 57.69 & 52.01 & 48.94 & 27.45 & 27.3 & 28.15 & 30.48 \\
        Wo-CR & 63.82 & 59.22 & 53.24 & 49.73 & 27.53 & 27.39 & 28.36 & 30.76 \\
        W-FT & 62.16 & 58.29 & 52.36 & 48.78 & 27.21 & 27.04 & 28.03 & 30.76 \\
        \bottomrule
    \end{tabular}
    \caption{Ablation experiments of QCG-rerank on Cultour and IIRC datasets.}
\end{table}

According to the results in Table 4,  Wo-CG represents QCG-rerank without chunks graph rerank, Wo-CR represents QCG-rerank without critical information extraction, and W-FT represents QCG-rerank only finetuned by the tourism dataset. Both components positively impact the final results. The critical information duplication method is confirmed to enhance the expressiveness of the semantic information in the query, thereby improving the overall performance. Additionally, the chunks graph rerank method effectively selects chunks with higher scores from the graph and inputs them into a fixed template. Subsequently, we utilize LLMs to summarize these chunks to answer the user query, thus enhancing the final results.
\subsection{Impact of fine-tuning embedding}
To evaluate the impact of fine-tuning embedding, we conducted tests using MRR@1, MRR@10, nDCG@1, and nDCG@10 to measure the accuracy of model retrieval after fine-tuning. As shown in Table 5, the experimental results demonstrate the improvements in fine-tuning the embedding model in MRR and nDCG metrics. These results indicate that fine-tuning the embedding model enhances the accuracy of retrieving domain-related documents, making the retrieved documents more relevant to the query. Consequently, it improves the overall performance of the model.
\begin{table}[h]
\centering
\scriptsize
\setlength{\tabcolsep}{2pt} % 减小列间距
\begin{tabular}{c|cccc|cccc}
\toprule
\multirow{2}{*}{Models}
& \multicolumn{4}{c}{Cultour} & \multicolumn{4}{c}{IIRC} \\
\cmidrule(lr){2-5} \cmidrule(lr){6-9}
& MRR@1 & MRR@10 & nDCG@1 & nDCG@10 & MRR@1 & MRR@10 & nDCG@1 & nDCG@10 \\ 
\midrule
Base & 57.69 & 67.24 & 57.63 & 71.37 & 10.79 & 18.02 & 10.42 & 22.14 \\
Base-finture & 60.49 & 68.76 & 60.45 & 72.62 & 19.73 & 28.00 & 18.67 & 31.99 \\
Large & 57.37 & 66.91 & 57.35 & 71.00 & 10.96 & 18.12 & 10.33 & 22.03 \\
Large-finture &\textbf{ 60.85} & \textbf{69.11} & \textbf{60.76} & \textbf{72.93} & \textbf{20.40} & \textbf{29.85} & \textbf{19.49 }& \textbf{34.28} \\
\bottomrule
\end{tabular}
\caption{Results of fine-tuning the embedding model on the Cultour and IIRC datasets.}
\label{tab:results}
\end{table}

\section{Discuss}
\subsection{impact of duplication times}
To further analyze the performance of QCG-rerank, we adjusted the number of times the critical information was duplicated. The specific results are shown in Table 6. On the Culture dataset, the best performance was achieved when the number of copies was set to 3, and on the IIRC dataset, the best performance was achieved when the number of copies was set to 2. This is because the critical information is enhanced in two aspects during the copying process: 1. The critical information in the query is emphasized as it is duplicated. 2. Since the critical information is duplicated, the query length becomes longer, and the length of the chunks in the database is relatively long, which makes it easier to increase the accuracy of matching, thereby improving the chunks retrieved and ultimately improving the effect of LLMs reply.

\begin{table}[h]
\centering
\scriptsize
\setlength{\tabcolsep}{4pt} % 减小列间距
\begin{tabular}{c|cccc|cccc}
\toprule
Numbers of  & \multicolumn{4}{c|}{Cultour} & \multicolumn{4}{c}{IIRC} \\
\cmidrule(r){2-5}\cmidrule(r){6-9}
         Critical        & R-1   & R-L   & B-1   & Met.     & R-1   & R-L   & B-1   & Met.     \\
\midrule
1                   & 64.02 & 59.54 & 53.45 & 49.98 & 27.99 & 27.80  & 29.24 & 31.66 \\
2                   & 64.49 & 59.92 & 54.04 & 50.35 & \textbf{28.85} & \textbf{28.66} & 29.61 & \textbf{32.32} \\
3                   & 64.69 & 60.12 & \textbf{54.41} & \textbf{50.49} & 27.51 & 27.27 & 28.66 & 31.18 \\
4                   & 64.60  & 59.98 & 54.25 & 50.53 & 28.25 & 28.01 & \textbf{29.74} & 31.56 \\
5                   & \textbf{64.71} & \textbf{60.13} & 54.32 & 50.39 & 28.01 & 27.77 & 29.35 & 31.45 \\
\bottomrule
\end{tabular}
\caption{The impact of duplication times.}
\label{table:model_results}
\end{table}

\subsection{impact of different amounts of content}
In addition, we evaluated the number of chunks input in the rerank phase, using chunks = \{5, 10, 15, 20\}. According to the results in Table 7, the best performance is achieved when the number of chunks is 10. As the number of chunks increases beyond 10, the model's performance does not significantly improve. However, due to the construction of the graph, the chunks graph rerank method increases the computational time as the number of chunks grows. Therefore, setting the number of chunks to 10 provides the best balance between performance and efficiency.
\begin{table}[h]
\centering
\scriptsize
\setlength{\tabcolsep}{4pt} % 减小列间距
\begin{tabular}{c|cccc|c}
\toprule
Numbers of 
 & \multicolumn{4}{c|}{Cultour} & \multirow{3}{*}{Time}\\
\cmidrule(r){2-5}
Contents & R-1 & R-L & B-1 & Met. & \\
\midrule
5     & 64.41 & 59.83 & 54.03 & 50.20 & \textbf{5,175} \\
10    & \textbf{64.69} & \textbf{60.12} & 54.41 & 50.49 & 5,310 \\
15    & 64.04 & 59.63 & 53.86 & 50.07 & 5,418 \\
20    & 64.67 & 60.08 & \textbf{54.43} & \textbf{50.50} & 5,528 \\
\bottomrule
\end{tabular}
\caption{The impact of different amounts of content in the rerank stage.}
\label{table:example}
\end{table}

\section{Conclusion}
Given that user queries in the tourism domain are typically brief, while the content in databases is often lengthy and complex, the retrieved information chunks following RAG may still contain numerous irrelevant or contradictory details. In this paper, we introduce a new model, QCG-rerank, which first uses tourism domain data to fine-tune the embedding model and improve the accuracy of data retrieval in the domain. Secondly, we use LLMs to extract critical information from the query and duplicate the critical information to update the query to enhance the matching between the updated query and the reference chunks in the rerank stage. Next, for the recall results, we construct a chunks graph and then use the chunks graph rerank method to select chunks with higher scores in the graph to input into the template. Subsequently, We use LLMs to summarize the input chunks to answer the query. We evaluate the model on Cultour, IIRC, StrategyQA, HotpotQA, SQuAD, and MuSiQue datasets. The experimental results demonstrate the effectiveness and superiority of the QCG-Rerank method. At the same time, we conduct ablation experiments to demonstrate the impact of each module on the model performance. In addition, we also profoundly analyzed the impact of the number of critical information duplications and the number of chunks input in the rerank phase on the model performance.

In future work, we will delve deeper into the unique characteristics of tourism data, extracting more in-depth features and integrating them into a knowledge graph. This integration aims to enhance the reasoning capabilities of our model, thereby enabling queries to match more relevant content and significantly improving the reliability and accuracy of the LLMs’ responses.

\bibliographystyle{elsarticle-num} 
\bibliography{mybibfile}

%% else use the following coding to input the bibitems directly in the
%% TeX file.

%% Refer following link for more details about bibliography and citations.
%% https://en.wikibooks.org/wiki/LaTeX/Bibliography_Management

%\begin{thebibliography}{00}

%% For numbered reference style
%% \bibitem{label}
%% Text of bibliographic item

%\bibitem{lamport94}
%  Leslie Lamport,
%  \textit{\LaTeX: a document preparation system},
%  Addison Wesley, Massachusetts,
%  2nd edition,
%  1994.

%\end{thebibliography}
\end{document}